\documentclass[lettersize,journal]{IEEEtran}
\usepackage{amsmath,amsfonts}
\usepackage{algorithmic}
\usepackage{algorithm}
\usepackage{array}
\usepackage[caption=false,font=normalsize,labelfont=sf,textfont=sf]{subfig}
\usepackage{textcomp}
\usepackage{stfloats}
\usepackage{url}
\usepackage{verbatim}
\usepackage{graphicx}
\usepackage{cite}
\usepackage{color}
\usepackage{multirow}
\usepackage{amsmath,amsfonts,amssymb}

\newcommand{\ve}[1]{\mathbf{#1}}

\usepackage{booktabs}

\begin{document}

\title{Classifier-head Informed Feature Masking and Prototype-based Logit Smoothing for Out-of-Distribution Detection}

\author{
Zhuohao Sun, Yiqiao Qiu, Zhijun Tan, Weishi Zheng, Ruixuan Wang

}



\maketitle

\begin{abstract}
Out-of-distribution (OOD) detection is essential when deploying neural networks in the real world. 
One main challenge is that neural networks often make overconfident predictions on OOD data. 
In this study, we propose an effective post-hoc OOD detection method based on a new feature masking strategy and a novel logit smoothing strategy.
Feature masking determines the important features at the penultimate layer for each in-distribution (ID) class based on the weights of the ID class in the classifier head and masks the rest features. Logit smoothing computes the cosine similarity between the feature vector of the test sample and the prototype of the predicted ID class at the penultimate layer and uses the similarity as an adaptive temperature factor on the logit to alleviate the network's overconfidence prediction for OOD data. 
With these strategies, we can reduce feature activation of OOD data and enlarge the gap in OOD score between ID and OOD data.
Extensive experiments on multiple standard OOD detection benchmarks demonstrate the effectiveness of our method and its compatibility with existing methods, with new state-of-the-art performance achieved from our method. The source code will be released publicly. 
\end{abstract}


\section{Introduction}


\IEEEPARstart{D}{eep} learning has made extraordinary achievements in various fields in recent years.
{ However, when deployed to the real world, deep learning models often encounter samples of unknown classes that were not seen during training~\cite{mos, MSP, 8640834}. 
These out-of-distribution (OOD) data may compromise the stability of the model, with potentially severe consequences such as in autonomous driving~\cite{autonomous,10227352,10168168}, medical diagnosis~\cite{medical}, and video anomaly detection~\cite{9266126,9828496}.} 
Therefore, deep learning models are expected to have the ability to detect whether any new data is OOD or not. 
There have been many explorations of OOD detection~\cite{reconstruction1,g-odin,huang2020feature,lin2021mood,ash,cider}. 
One line of work is post-hoc, i.e., the model is pre-trained and fixed and the focus is on how to design an effective scoring function~\cite{MSP,energy,featurenorm} to measure the degree of any new input belonging to one of the learned classes. Any data from any learned class is called in-distribution (ID). The aim is to assign higher scores to ID data and lower scores to OOD data.
Such post-hoc methods have practical significance when deploying models to the real world as it does not require any additional design of new training modules for OOD detection. 
However, recent studies reveal that neural networks have overconfident predictions for OOD inputs~\cite{overconfidence,relu}, resulting in small difference in score between ID and OOD data. 
Therefore, how to solve the overconfidence problem of neural networks and make the activation of networks as small as possible for OOD data is the key to improving OOD detection performance.

\begin{figure}[t]
\centering
\includegraphics[width=0.5\textwidth]{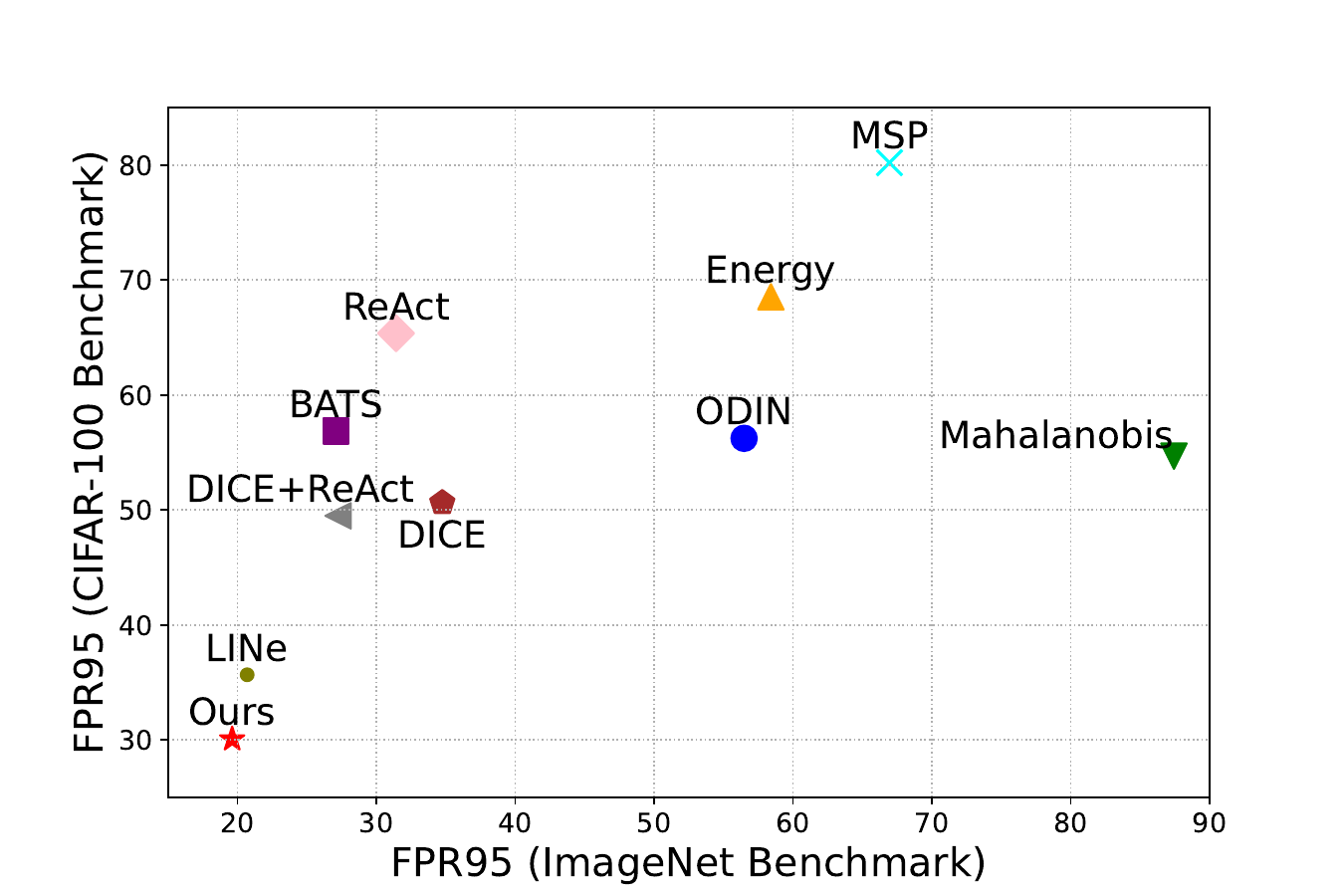} 
\caption{OOD detection performance from different methods on CIFAR-100 and ImageNet benchmarks. Smaller FPR95 values mean better performance. }
\label{performance}
\end{figure}

\begin{figure}[t]
\centering
\includegraphics[width=0.47\textwidth]{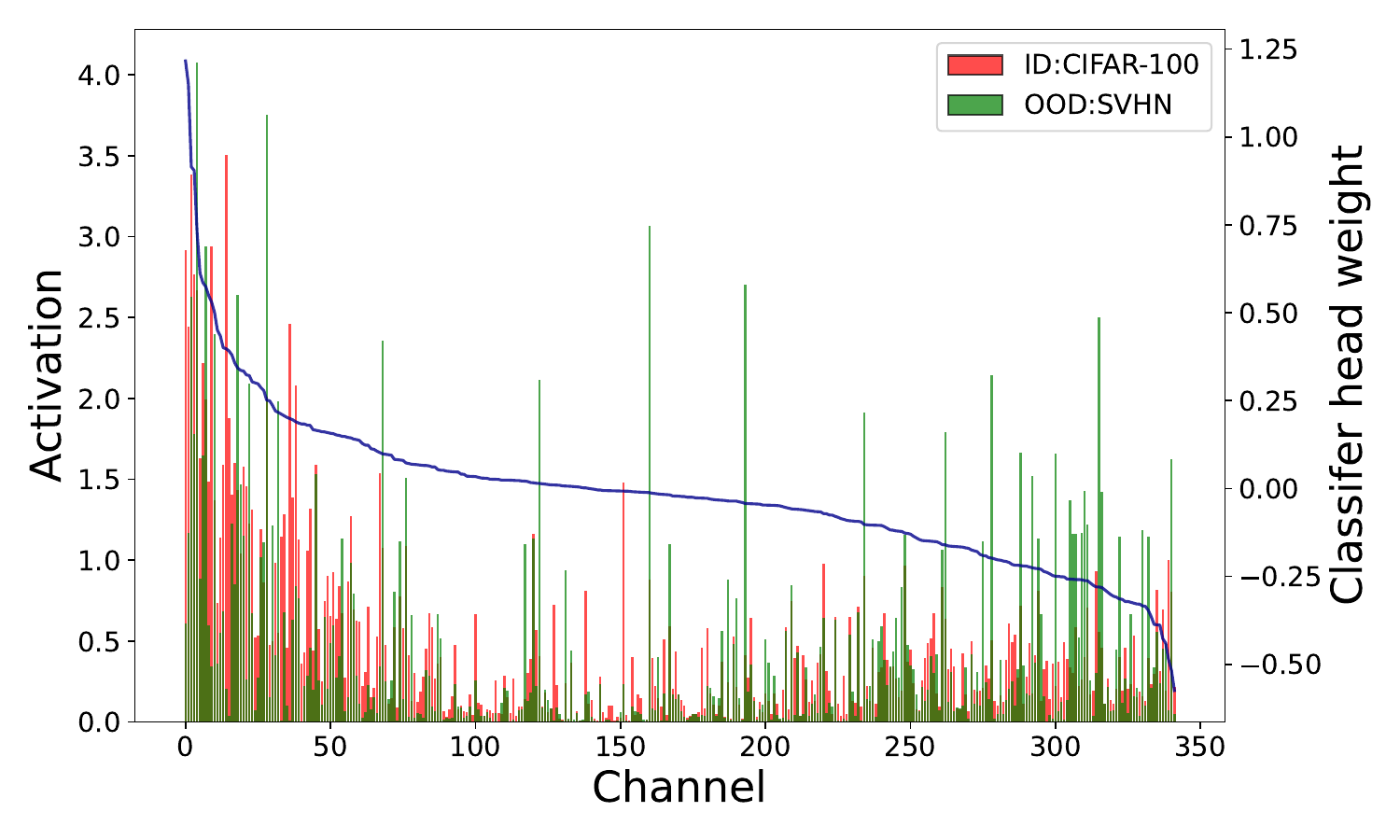} 
\caption{Distribution of weights (blue curve) in the classifier head associated with a specific ID class, and distribution of activation over feature channels at the penultimate layer from ID (red histogram) and OOD data (green histogram). 
Activation of each feature is the average over the data of the same ID class in CIFAR-100 or over the OOD data which are predicted as the ID class from the OOD dataset SVHN. The feature channels are sorted in descending order by the weights of the ID class in the classifier head. Decreasing weights from the left to the right is largely correlated with the overall decreasing trend of feature activation from ID data, but not from OOD data.}
\label{intro}
\end{figure}
\begin{figure}[t]
\centering
\includegraphics[width=0.48\textwidth]{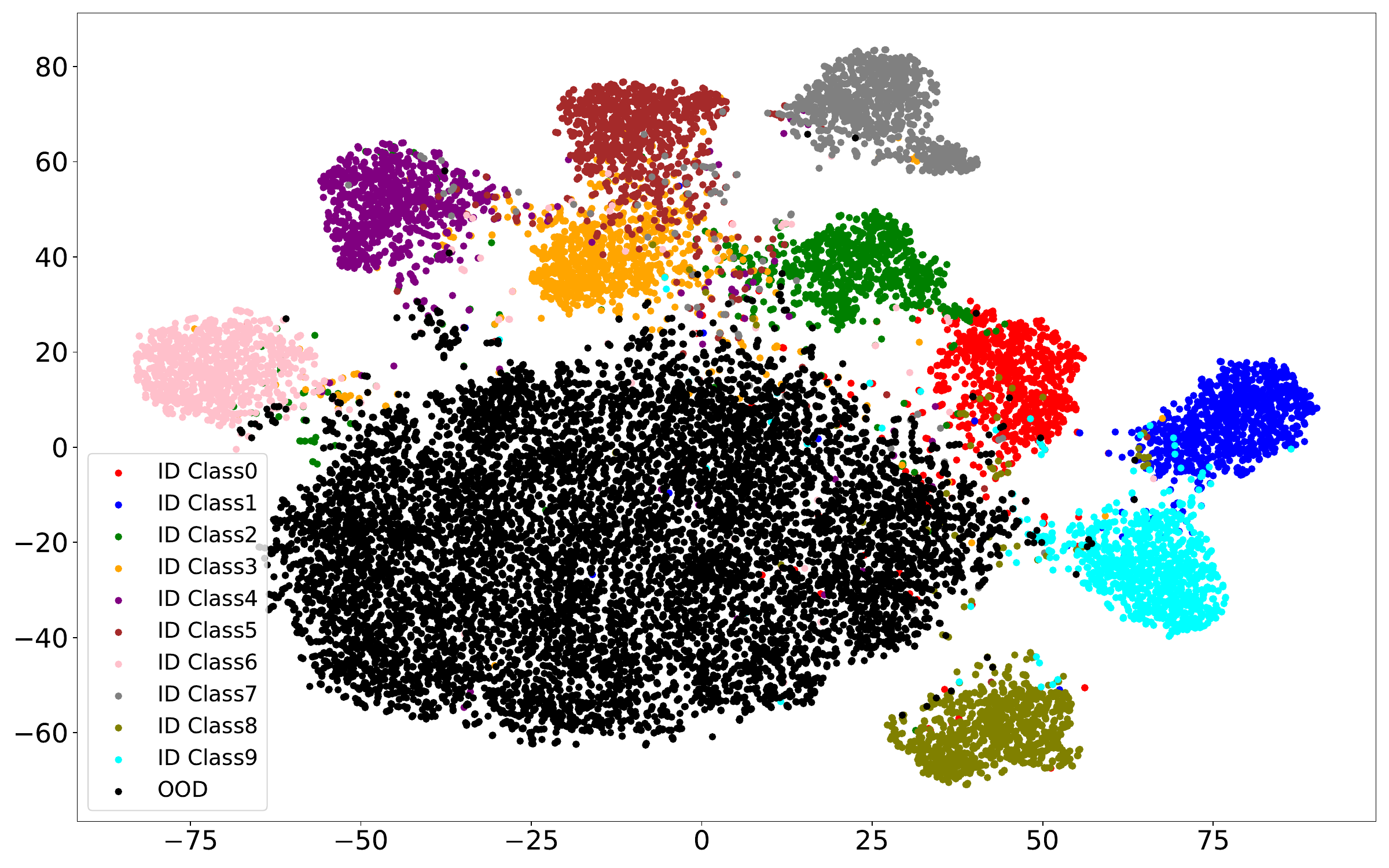} 
\caption{Visualization of the distribution of ID and OOD feature vectors at the penultimate layer by t-SNE~\cite{tsne}. The model is DenseNet~\cite{densenet} trained on the CIFAR-10 dataset. The ID dataset is CIFAR-10 and OOD dataset is LSUN. Overall OOD data are far away from the any ID class.}
\label{t-sne}
\end{figure}

This study provides two new strategies to reduce the feature activation of OOD data, while preserving the activation of ID data largely unchanged, thus improving the separability in detection score between ID and OOD data. The first strategy, called feature masking, is inspired by the observation that the feature activation at the penultimate layer is positively correlated with the associated weights in the classifier head for each class of ID data, while not for OOD data (Figure~\ref{intro}). This observation is consistent with the efficacy of widely used model interpretation methods CAM~\cite{cam} and Grad-CAM~\cite{gradcam} which use the weights in the classifier head to represent the importance of feature elements when interpreting the predicted specific class. Based on this observation, we propose to determine the important features at the penultimate layer for each ID class according to the weights in the classifier head associated with the ID class, and then mask the other features which are less important to the ID class. 
In doing so, most of the high activation units that play an important role in classification can be preserved for ID data, while those high feature activation appearing in the removed units from OOD data are removed, thus largely reducing the feature activation of OOD data.
{The second strategy, called logit smoothing, is motivated by the observed difference in feature vector distribution between ID and OOD data at the penultimate layer as shown in Figure~\ref{t-sne}. }In particular, ID data is often close to its class center (`prototype') while OOD data is relatively not close to any ID class center in the feature space. With such observed difference, the cosine similarity between new input data and the prototype of the predicted ID class at the penultimate layer is used to tune the logit vector in the output layer. Such a combination further enlarges the OOD score gap between ID and OOD data. 
In summary, the main contributions are as follows:
\begin{itemize}
    \item We propose a new post-hoc OOD detection method that reduces feature activation of OOD data to enlarge the gap in OOD score between ID and OOD data. 
    \item We introduce a new class-specific feature masking strategy simply based on weights in the classifier head.
    \item We propose a novel logit smoothing strategy combining information at the penultimate layer with logits for better OOD detection.
     
    \item We extensively evaluate our method on multiple standard benchmarks and show the compatibility with existing methods, with new state-of-the-art performance achieved. 
    

\end{itemize}

\section{Related Work}
The key to OOD detection is to find potentially different output patterns from the model for ID and OOD data. Reconstruction-based methods show that an encoder-decoder framework trained on ID data usually produces different quantities of reconstruction errors for ID and OOD samples~\cite{yang2022out, reconstruction3,reconstruction2,9862976}. 
In particular, a reconstruction model trained only on ID data cannot recover OOD data well. Distance-based methods measure the distance between the input sample and the centroid or prototype of ID class in the feature space to detect OOD samples~\cite{tack2020csi,ssd}. For example, Mahalanobis score~\cite{mahala} uses the Mahalanobis distance between the feature vector of input sample and the prototype feature vector of training data for OOD detection. KNN~\cite{knn} computes the $k$-th nearest neighbor distance between the feature vector of input sample and the feature vector of the training set. 
Besides, confidence enhancement methods attempt to enhance the confidence of the network via data augmentation~\cite{10106029} or designing a confidence estimation branch etc.~\cite{devries2018learning, relu, vyas2018out}.
{~\cite{10106029} mixed ID samples to generate fake OOD samples for enhancing the confidence of ID samples.}
LogitNorm~\cite{logitnorm} proposes to enforce a constant norm on the logit during training to alleviate the overconfidence of the neural network.

Differently, post-hoc methods attempt to perform OOD detection by designing a scoring function for a pre-trained and fixed model, assigning an OOD score to each new  input~\cite{gradnorm,wang2022vim,SHE}. For example, MSP~\cite{MSP} provides a simple baseline for OOD detection by using the maximum probability output of the model. ODIN~\cite{ODIN} introduces two operations based on MSP called temperature scaling and input perturbation to separate OOD from ID samples. Energy score~\cite{energy} 
uses the energy function of the logits (i.e., input to the softmax at the output layer) for OOD detection. 
To alleviate the overconfidence problem of the model on OOD data, based on the observation that OOD data often cause abnormally high activation at the penultimate layer of the network, ReAct~\cite{react} rectifies feature activation at an upper limit and reduces most of the activation values caused by OOD data. 
DICE~\cite{dice} reduces the variance of the output distribution by clipping some noise units irrelevant to ID classes, resulting in improved separability in the OOD score distribution between ID and OOD data. 
LINe~\cite{LINe} employs the Shapley value~\cite{shapley} to measure each neuron’s contribution and reduces the effect of less important neurons at the last network layer. 

Our method is similar to LINe and DICE, as ours and the two methods all propose a sparsification strategy for the model to reduce feature activation of OOD data. While both LINe and DICE need a complicated process to select important elements using a training dataset, our method simply utilizes the classifier head's weight to select the important feature elements for each ID class. In addition, our method uses a novel logit smoothing operator to combine the feature information at the penultimate layer and the logit information to further improve OOD detection. 


\begin{figure*}[t]
\centering
\includegraphics[width=0.97\textwidth]{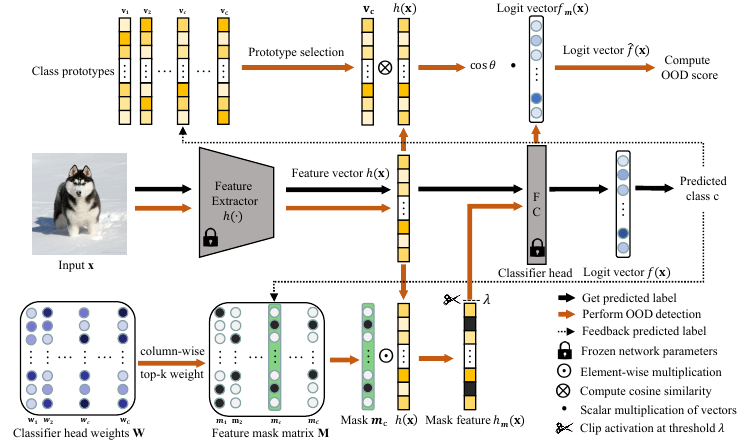} 
\caption{Overview of the proposed framework. Given a well-trained classifier (middle row) and any test data $\ve{x}$, the post-hoc feature masking (bottom row) and logit smoothing (top row) together modulate the original logit vector $f(\ve{x})$, and the OOD score based on the modulated logit vector $\hat{f}(\ve{x})$ is computed (top right) for OOD detection. 
}
\label{method}
\end{figure*}

\section{Preliminary}
Consider a neural network classifier trained with a training set $\mathcal{D} = \{ (\ve{x}_i, y_i)\}^N_{i=1}$, where $\ve{x}_i$ is the $i$-th training image and $y_i \in \{1, 2, \ldots, C\}$ is the associated class label. When deploying the neural network classifier in the real world, new data may be from a certain unknown distribution which is different from the distribution of the training data. 
Such data are out-of-distribution (OOD) and should not be predicted as any of the in-distribution (ID) classes learned during classifier training. 
The task of OOD detection is to identify whether any new data is ID or OOD. 




OOD detection can be considered as a binary classification problem. In particular, 
a scoring function $S(\ve{x}; f)$ can be designed to estimate the degree of any new data $\ve{x}$ belonging to any of the ID classes, where the function $f$ denotes the overall feature extraction process whose output is used as the input to the scoring function. With the scoring function, OOD detection can be simply formulated as a binary classifier $g (\ve{x}; f)$ as below,
\begin{equation}
g (\ve{x}; f)=
\begin{cases}
\text{1} & \text{ if } S (\ve{x}; f) \ge \gamma  \\
\text{0} & \text{ if } S (\ve{x}; f) < \gamma 
\end{cases},
\end{equation}
where data with higher scores $S (\mathrm{x}; f)$ are classified as ID (with label 1) and lower scores are classified as OOD (with label 0), and $\gamma$ is the threshold hyperparameter.



\section{Method}


An overview of the proposed post-hoc OOD detection framework is illustrated in Figure~\ref{method}. Given a pre-trained neural network (Figure~\ref{method}, components in gray) in this framework, the feature elements at the penultimate layer which are more relevant to each ID class can be identified based on the weight parameters of the classifier head,  
and then feature masking is performed by masking the less important feature elements. On the other hand, motivated by the observed differences between OOD and ID samples in the feature space, logit smoothing is applied by combining information from the penultimate layer's features and the output layer's logit before estimating the OOD score. 
In addition, as in the state-of-the-art method LINe~\cite{LINe}, the clipping of feature activation at the penultimate layer called ReAct~\cite{react} is also applied here.

\begin{figure*}[!t]
\centering
\subfloat[iNaturalist]{\includegraphics[width=0.22\textwidth]{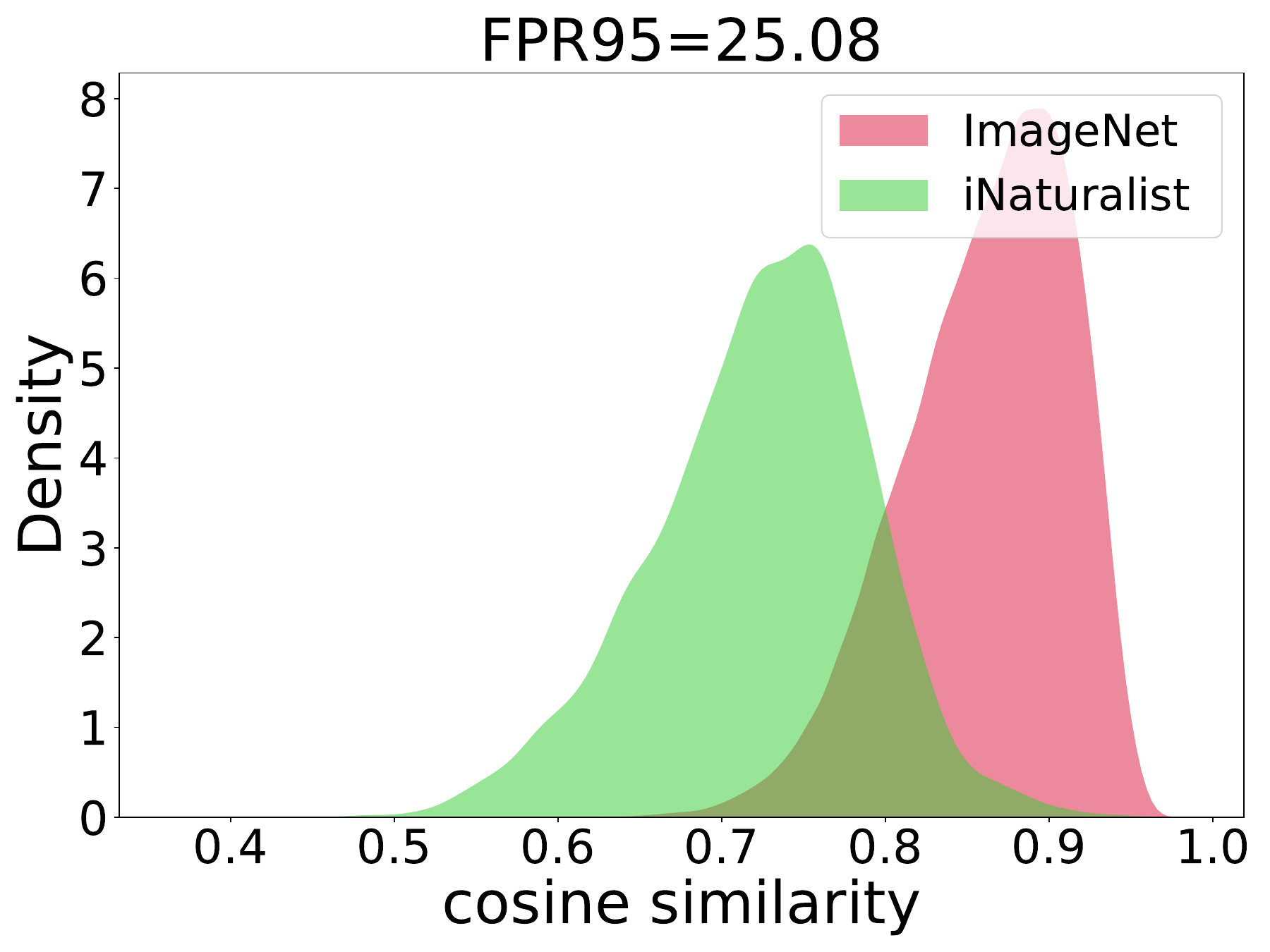}%
\label{fig:cosine_inat}}
\hfil
\subfloat[SUN]{\includegraphics[width=0.22\textwidth]{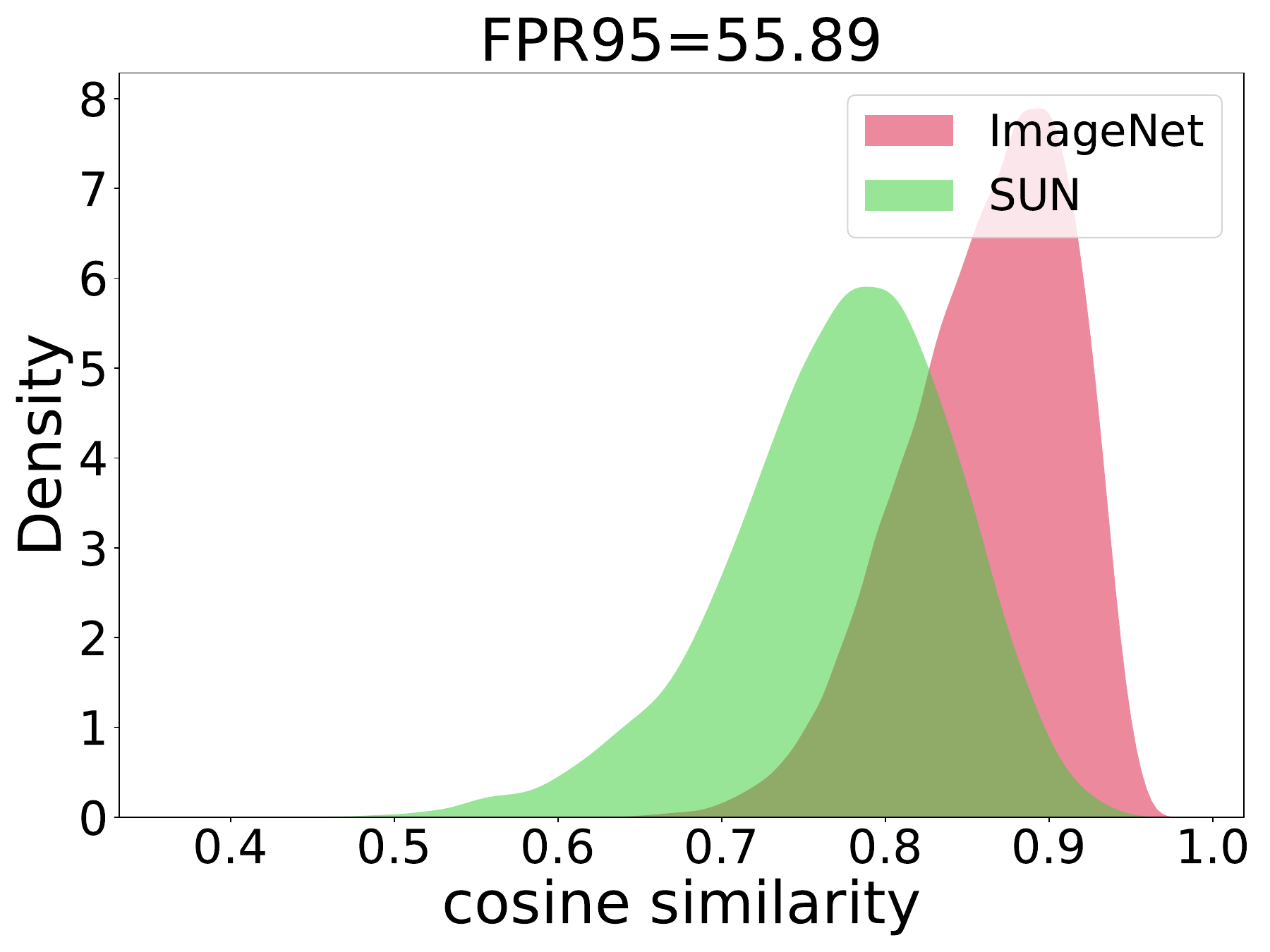}%
\label{fig:cosine_sun}}
\hfil
\subfloat[Places]{\includegraphics[width=0.22\textwidth]{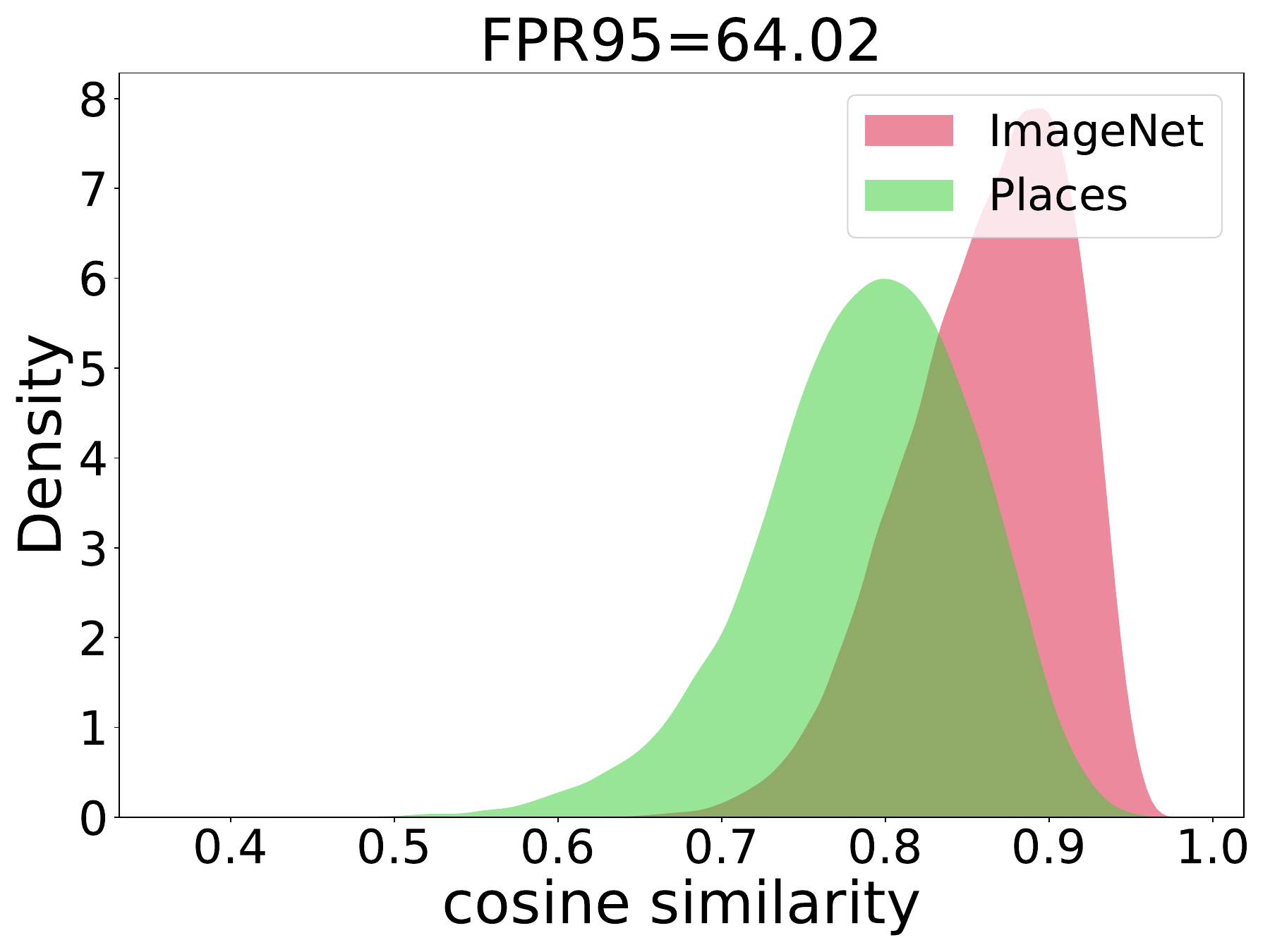}%
\label{fig:cosine_places}}
\hfil
\subfloat[Textures]{\includegraphics[width=0.22\textwidth]{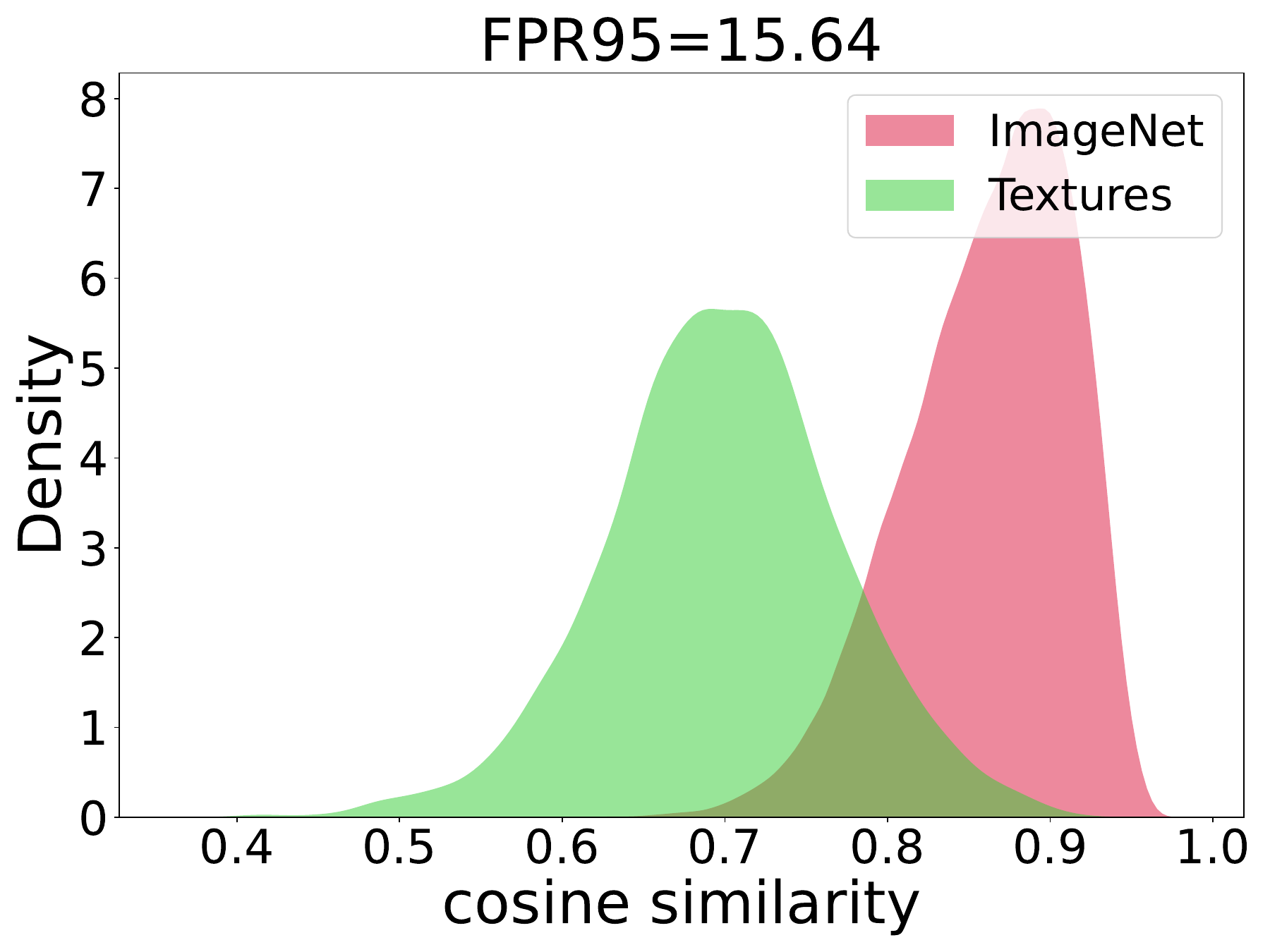}%
\label{fig:cosine_textures}}
\caption{Distribution of the cosine similarity between the feature of ID/OOD data and the feature of the class prototype on ResNet-50. ID data refers to ImageNet-1k and FPR95 is calculated based on the cosine similarity for ID and OOD data.}

\label{fig:cosine}
\end{figure*}

\subsection{Feature Masking}
For a well-trained network, given any test data $\ve{x}$,  
denote by $h (\ve{x}) \in \mathbb{R}^L$ the feature vector from the penultimate layer of the network. The classifier head's weight matrix $\ve{W} \in \mathbb{R}^{L \times C}$ together with the bias vector $\ve{b}$ transforms the feature vector $h (\ve{x})$ to the output logit vector $f(\ve{x})$ as follows,
\begin{equation}
f(\ve{x}) = \ve{W}^{\intercal} h(\ve{x}) + \ve{b} \,.
\end{equation}
Inspired by the observation in Figure~\ref{intro} and the model output interpretation methods CAM~\cite{cam} and Grad-CAM~\cite{gradcam}, where the weights in the classifier head are correlated with the importance of feature channels from the penultimate layer for each class, we propose selecting the top-$k$ weights for each class based on the $k$-largest elements from each column in $\mathbf{W}$. 
Specifically, denote by $\textbf{M} \in \mathbb{R}^{L \times C}$ the binary feature mask matrix, where 1 is set for the $k$-largest elements from each column in $\mathbf{W}$ and 0 for the rest of the column, and suppose the neural network classifier predicts the new data $\ve{x}$ as class $c$, then the modulated feature vector by the feature mask of class $c$ from the penultimate layer becomes
\begin{equation} \label{eq:mask}
h_m(\ve{x}) = \ve{m}_c  \odot h(\ve{x}) \,,
\end{equation}
where $\ve{m}_c \in \mathbb{R}^L$ is the $c$-th column of $\ve{M}$ representing the feature mask of class $c$, and $\odot$ represents the element-wise multiplication. In the mask-modulated feature vector $h_m(\ve{x})$, those feature elements whose activation is masked (i.e., removed)  often have smaller activation values and therefore have little effect on the prediction of class $c$. If the input $\ve{x}$ does belong to ID class $c$, the modulated feature vector $h_m(\ve{x})$ would be still quite similar to the original feature vector $h(\ve{x})$, and therefore the modulation operator based on the feature mask $\ve{m}_c$ would not likely change the original prediction for the input $\ve{x}$. In contrast, if the input $\ve{x}$ is an OOD data, and it is originally misclassified as ID class $c$, such misclassification may be partly from relatively stronger activation of the to-be-masked feature elements. In this case, the modulation based on the feature mask $\ve{m}_c$ would result in a modulated feature vector $h_m(\ve{x})$ whose overall activation (i.e., the norm of feature vector) would be substantially reduced. Considering that feature activation of ID data is often statistically stronger than that of OOD data as observed in previous studies~\cite{react,featurenorm}, and that scoring functions are often designed by utilizing such difference in feature activation between ID and OOD data~\cite{energy,ash}, the potentially substantial reduction in feature activation for OOD data in the modulated feature vector $h_m(\ve{x})$ would further enlarge the gap in feature activation between ID and OOD data, making OOD detection easier. 

In addition, on a similar rationale of further reducing feature activation of OOD data, following the previous studies~\cite{react,LINe}, the clipping operator ReAct~\cite{react} is applied to the masked feature $h_m (\mathrm{x})$ to reduce substantially higher feature activation which often appears only in OOD, i.e., 
\begin{equation} \label{eq:react}
\overline{h}_m (\ve{x}) = \texttt{ReAct} (h_m(\ve{x}); \lambda),
\end{equation}
where $\texttt{ReAct}(\ve{x}; \lambda)=\min(\mathrm{x}, \lambda)$ which is applied element-wise to $h_m(\ve{x})$, and $\lambda$ is a threshold hyperparameter.

With the mask modulation and the ReAct clipping operators, the output logit vector becomes 
\begin{equation}
f_{m}(\ve{x}) = \ve{W}^{\intercal} \overline{h}_m (\ve{x}) + \textbf{b} \,.
\end{equation}

\subsection{Logit Smoothing}
For any test data $\ve{x}$, still from its originally predicted class $c$, the cosine similarity between the feature vector $h(\ve{x})$ of the test data and the prototype of class $c$ in the penultimate layer's feature space can be utilized to help separate OOD data from ID data. 
{Denote by $\ve{v}_c \in \mathbb{R}^{L}$ the prototype of class $c$ which can be estimated in advance by the average of feature vectors over all the training data belonging to class $c$, i.e,
\begin{equation}
\textbf{v}_c = \frac{1}{N_c}\sum_{i:y_i=c}h (\mathrm{x}_i),
\end{equation}
where $N_c$ is the number of the training data belonging to class $c$.

Simultaneously, we denote by $s(h(\ve{x}), \ve{v}_c)$ the cosine similarity between $h(\ve{x})$ and $\ve{v}_c$:
\begin{equation}
s(h(\ve{x}), \ve{v}_c) = \cos (\theta) = \frac{{h(\ve{x}) \cdot \ve{v}_c}}{{\|h(\ve{x})\| \|\ve{v}_c\|}}.
\end{equation}
}
Then the logit vector $f(\ve{x})$ can be modulated by the cosine similarity $s(h(\ve{x}), \ve{v}_c)$ as follows,
\begin{equation}
f_{s}(\ve{x}) = s(h(\ve{x}), \ve{v}_c) \cdot f(\ve{x}) \,.
\end{equation}
If the input $\ve{x}$ is indeed from ID class $c$, it is expected that the cosine similarity $s(h(\ve{x}), \ve{v}_c)$ between the feature vector of the input and its class prototype in general is higher. In contrast, if $\ve{x}$ is an OOD data, its feature vector $h(\ve{x})$ is in general not within the distribution of class $c$ in the feature space, and therefore the cosine similarity would be lower. As shown in Figure~\ref{fig:cosine}, we can clearly observe this phenomenon. Such differences between ID and OOD data in cosine similarity can be utilized to help design a more effective scoring function for OOD detection (see Equation~\ref{eq:energy} below).

The cosine similarity $s(h(\ve{x}), \ve{v}_c)$ in the modulated logit vector $f_{s}(\ve{x})$ can be viewed as an input-adaptive temperature $\tau$ on the logit vector $f(\ve{x})$, specifically by setting $\tau = 1/s(h(\ve{x}), \ve{v}_c)$. With the temperature-tuned logits, the output of class $c$ after the softmax operator becomes
\begin{equation}
\text{softmax}_c(f(\ve{x}); \tau) = \frac{\exp(f_c(\ve{x})/\tau)}{\sum\limits_{k=1}\limits^{C} \exp(f_k(\ve{x})/\tau)} \,,
\end{equation}
where $f_c(\ve{x})$ is the logit element of the predicted class $c$ in the logit vector$f(\ve{x})$. Since the cosine similarity $s(h(\ve{x}), \ve{v}_c)$ is in general smaller for OOD data than for ID data, $\tau$ would be larger for OOD data. As a result, the softmax outputs become smoother (i.e., closer to a discrete uniform distribution) for OOD data, and correspondingly the confidence of predicting the OOD data as ID class $c$ becomes lower. Since the modulated logit vector $f_{s}(\ve{x})$ can alleviate overconfidence of predicting OOD data as ID classes, it can be also applied to softmax-based OOD detection methods such as MSP~\cite{MSP} and ODIN~\cite{ODIN}. Considering the smoothing effect on the softmax outputs, the modulation of the logit vector by the cosine similarity is called logit smoothing.

\subsection{Scoring Function}

Combining feature masking and logit smoothing, the modulated final logit vector is
\begin{eqnarray}
    \hat{f}(\ve{x}) & = & s(h(\ve{x}), \ve{v}_c) \cdot f_m(\ve{x}) \nonumber \\
    & = & s(h(\ve{x}), \ve{v}_c) \cdot \big\{\ve{W}^{\intercal} \overline{h}_m (\ve{x}) + \textbf{b} \big\} \,.
\end{eqnarray}
While various scoring functions can be applied based on $\hat{f}(\ve{x})$, the energy scoring function~\cite{energy} is used by default, i.e.,
\begin{equation}
    S(\ve{x}; \hat{f}) = \log \sum_{k=1}^{C}\exp(\hat{f}_k (\ve{x})) \,,
\label{eq:energy}
\end{equation}
where $\hat{f}_k (\ve{x})$ is the  $k$-th element in the modulated final logit vector $\hat{f}(\ve{x})$. 
Since the cosine similarity $ s(h(\ve{x}), \ve{v}_c)$ is positively correlated with the energy score $S(\ve{x}; \hat{f})$, lower cosine similarity from OOD data will lead to smaller energy score, while higher cosine similarity from ID data will lead to larger energy score. Similarly for the modulated logit vector $f_m(\ve{x})$ by the mask modulation and the ReAct clipping operators (Equations~\ref{eq:mask} and~\ref{eq:react}). Overall, feature masking and logit smoothing help enlarge the gap in energy score between ID and OOD data.

\subsection{Differences from LINe}
Our method is partly inspired by the state-of-the-art method LINe~\cite{LINe} in masking features based on estimated important feature elements for the predicted class. However, our method is significantly different from LINe in multiple aspects.
First, the masking strategy is different and ours is much simpler and more efficient. LINe employs the Shapley value~\cite{shapley} to measure each feature's contribution for each class, and generates masks based on these values. Notably, LINe needs training data to calculate the Shapley value. In contrast, our method simply uses the weight parameters of the classifier head to select the important features for each class. 
Second, 
our method includes a novel logit smoothing operator 
which can further enlarge the difference in the modulated logit vector between ID and OOD data and also alleviate the overconfidence of OOD predictions. 
Third, our method outperforms LINe on standard benchmarks with different model backbones. 
Furthermore, our method is compatible with other OOD detection methods, and even the performance of LINe can be further improved when combined with the proposed logit smoothing.

{Noticeably, compared to other existing methods~\cite{dice, react, logitnorm} for alleviating the overconfidence problem of the model on OOD data, our method offers several compelling advantages:~(1) We can remove most of the high activation feature of OOD samples without compromising the accuracy of ID classification by simply utilizing the information of classifier head’s weight.~(2) We utilize the feature information at the penultimate layer as an input-adaptive temperature on the logit vector during inference time which can further mitigate overconfidence of predicting OOD data as ID classes.~(3) Our method can be efficiently applied to various model architectures and is robust to the choice of hyperparameters.}

\section{Experiments}

\begin{table*}[!tbh]
\begin{center}
\caption{Comparison between different methods in OOD detection on the CIFAR-10 and CIFAR-100 benchmarks with different model backbones. $\downarrow$ indicates smaller values mean better performance and $\uparrow$ indicates larger values mean better performance. Bold numbers are superior results and underlined numbers are the 2nd best results. All values are percentages.}
\label{tab:cifar benchmark}
\resizebox{\textwidth}{!}{%
\begin{tabular}{cccccccccccccccc}
\toprule
\multicolumn{1}{c}{\multirow{3}{*}{\begin{tabular}[c]{@{}c@{}}ID Dataset\\ Model\end{tabular}}} & \multicolumn{1}{c}{\multirow{3}{*}{Method}} & \multicolumn{12}{c}{OOD Datasets} & \multicolumn{2}{c}{\multirow{2}{*}{Average}} \\ \cline{3-14} 
 \multicolumn{1}{c}{} & \multicolumn{1}{c}{} & \multicolumn{2}{c}{SVHN} & \multicolumn{2}{c}{LSUN-C} & \multicolumn{2}{c}{LSUN-R} & \multicolumn{2}{c}{iSUN} & \multicolumn{2}{c}{Textures} & \multicolumn{2}{c}{Places365} \\ 
 &  & FPR95$\downarrow$ & AUROC$\uparrow$ & FPR95$\downarrow$ & AUROC$\uparrow$ & FPR95$\downarrow$ & AUROC$\uparrow$ & FPR95$\downarrow$ & AUROC$\uparrow$ & FPR95$\downarrow$ & AUROC$\uparrow$ & FPR95$\downarrow$ & AUROC$\uparrow$ & FPR95$\downarrow$ & AUROC$\uparrow$ \\ \midrule
\multirow{10}{*}{\begin{tabular}[c]{@{}c@{}}CIFAR-10 \\ResNet-18\end{tabular}} 
& MSP~\cite{MSP} & 57.52 & 92.05 & 43.53 & 94.05 & 55.65 & 91.88 & 56.10 & 91.74 & 61.58 & 89.60 & 61.28 & 88.96 & 55.94 & 91.38 \\
& ODIN~\cite{ODIN} & 35.06 & 94.00 & 11.37 & 97.89 & \underline{25.82} & \textbf{95.58} & \underline{27.53} & \textbf{95.26} & 47.57 & 89.86 & 42.28 & 90.66 & \underline{31.60} & 93.87 \\
& Mahalanobis~\cite{mahala} & 26.10 & 93.74 & 47.22 & 77.33 & \textbf{10.98} & 90.73 & \textbf{11.80} & 91.26 & \textbf{37.50} & 90.38 & 86.54 & 64.60 & 36.69 & 84.67 \\
& Energy~\cite{energy} & 36.14 & 93.95 & 11.97 & 97.80 & 29.40 & 94.98 & 31.26 & 94.66 & 48.62 & 89.72 & 42.33 & 90.75 & 33.29 & 93.64 \\
& BATS~\cite{bats} & 35.81 & 93.99 & 11.87 & 97.82 & 28.37 & 95.19 & 30.13 & 94.88 & 47.96 & 90.24 & \underline{42.23} & 90.76 & 32.73 & 93.81 \\
& DICE~\cite{dice} & 34.04 & 94.04 & 8.89 & 98.43 & 33.15 & 94.55 & 35.57 & 94.03 & 52.22 & 88.56 & 44.71 & 90.30 & 34.76 & 93.32 \\
& ReAct~\cite{react} & 40.06 & 93.65 & 15.02 & 97.48 & 28.41 & \underline{95.22} & 30.25 & \underline{94.98} & 45.25 & 91.36 & \textbf{40.98} & \textbf{91.60} & 33.33 & \underline{94.05} \\
& DICE+ReAct~\cite{dice} & 36.39 & 93.74 & 9.86 & 98.27 & 31.58 & 94.75 & 34.01 & 94.31 & 48.37 & 90.43 & 43.04 & \underline{91.03} & 33.87 & 93.75 \\
& LINe~\cite{LINe} & \textbf{18.63} & \textbf{96.94} & \underline{8.41} & \underline{98.46} & 42.01 & 93.31 & 45.14 & 92.61 & 42.77 & \underline{92.10} & 62.44 & 84.67 & 36.57 & 93.01 \\
& Ours & \underline{22.30} & \underline{96.26} & \textbf{7.17} & \textbf{98.70} & 31.65 & 94.79 & 35.15 & 94.19 & \underline{39.04} & \textbf{92.79} & 50.26 & 89.17 & \textbf{30.92} & \textbf{94.32} \\

\midrule

\multirow{10}{*}{\begin{tabular}[c]{@{}c@{}}CIFAR-10  \\DenseNet\end{tabular}} 
& MSP~\cite{MSP} & 60.03 & 90.84 & 32.25 & 95.83 & 39.52 & 94.69 & 42.58 & 94.24 & 63.48 & 87.94 & 62.41 & 88.74 & 50.04 & 92.05 \\
& ODIN~\cite{ODIN} & 14.52 & 97.14 & 2.98 & 99.25 & \textbf{3.39} & \textbf{99.02} & \textbf{4.64} & \textbf{98.83} & 57.85 & 81.18 & 49.44 & 89.60 & 22.14 & 94.17 \\
& Mahalanobis~\cite{mahala} & 35.28 & 88.44 & 18.24 & 93.18 & 7.34 & 97.88 & 8.72 & 97.76 & 30.41 & 89.53 & 87.52 & 64.83 & 31.25 & 88.60 \\
& Energy~\cite{energy} & 42.73 & 93.27 & 3.46 & 99.18 & 9.44 & 98.21 & 12.49 & 97.89 & 59.20 & 85.75 & 42.35 & 91.55 & 28.28 & 94.31 \\
& BATS~\cite{bats} & 15.42 & 97.35 & 3.16 & 99.25 & 8.17 & 98.38 & 10.22 & 98.05 & 45.80 & 92.03 & \textbf{41.04} & \underline{91.57} & 20.63 & 96.10 \\
& DICE~\cite{dice} & 34.72 & 93.61 & \underline{0.39} & \underline{99.89} & \underline{5.48} & \underline{98.85} & \underline{6.33} & \underline{98.75} & 46.35 & 87.33 & 45.65 & 90.25 & 23.15 & 94.78 \\
& ReAct~\cite{react} & 18.94 & 96.84 & 4.33 & 99.05 & 10.78 & 97.99 & 13.70 & 97.57 & 51.52 & 90.19 & \underline{41.53} & \textbf{91.60} & 23.47 & 95.54 \\
& DICE+ReAct~\cite{dice} & \underline{4.98} & \underline{98.89} & \textbf{0.27} & \textbf{99.90} & 8.43 & 98.33 & 9.19 & 98.22 & \underline{28.30} & 94.30 & 51.50 & 88.70 & 17.11 & 96.39 \\
& LINe~\cite{LINe} & 6.18 & 98.78 & 0.48 & 99.84 & 8.87 & 98.34 & 10.17 & 98.20 & 28.56 & \underline{94.53} & 47.44 & 89.86 & \underline{16.95} & \underline{96.59} \\
& \textbf{Ours} & \textbf{4.19} & \textbf{99.12} & 0.56 & 99.83 & 6.71 & 98.79 & 7.63 & 98.62 & \textbf{25.18} & \textbf{95.53} & 43.42 & 91.19 & \textbf{14.62} & \textbf{97.18} \\

\midrule
\multirow{10}{*}{\begin{tabular}[c]{@{}c@{}}CIFAR-100  \\ResNet-18\end{tabular}} 
& MSP~\cite{MSP} & 83.67 & 75.41 & 72.36 & 83.31 & 81.84 & 75.21 & 83.39 & 73.90 & 85.62 & 73.58 & 82.78 & 74.97 & 81.61 & 76.06 \\
& ODIN~\cite{ODIN} & 85.97 & 80.86 & 37.97 & 93.72 & 65.83 & 85.35 & 69.39 & 83.52 & 83.85 & 75.53 & 80.10 & 77.53 & 70.52 & 82.75 \\
& Mahalanobis~\cite{mahala} & 58.59 & 75.92 & 84.92 & 49.51 & 59.04 & 61.12 & \underline{52.25} & 67.38 & 76.31 & 47.72 & 93.68 & 58.13 & 70.80 & 59.96 \\
& Energy~\cite{energy} & 84.07 & 81.97 & 35.48 & 94.01 & 68.88 & 84.19 & 72.29 & 82.24 & 84.66 & 75.14 & 80.28 & 77.39 & 70.95 & 82.49 \\
& BATS~\cite{bats} & 68.28 & 88.67 & 32.67 & 94.00 & \underline{56.96} & \underline{88.10} & 59.45 & \underline{87.28} & 55.12 & 89.16 & 77.91 & \underline{78.33} & 58.40 & 87.59 \\
& DICE~\cite{dice} & 64.56 & 87.52 & \textbf{7.14} & \textbf{98.63} & 65.24 & 85.47 & 66.48 & 84.41 & 74.59 & 76.54 & 78.22 & 77.93 & 59.37 & 85.08 \\
& ReAct~\cite{react} & 72.27 & 87.04 & 41.37 & 92.15 & \textbf{49.46} & \textbf{89.98} & \textbf{52.04} & \textbf{89.22} & 51.21 & 89.18 & \textbf{75.53} & \textbf{79.41} & 56.98 & 87.83 \\
& DICE+ReAct~\cite{dice} & \underline{39.63} & \underline{93.04} & \underline{7.99} & \underline{98.32} & 74.78 & 85.52 & 71.79 & 86.64 & 45.64 & 88.81 & 82.29 & 75.44 & 53.69 & 87.96 \\
& LINe~\cite{LINe} & 43.26 & 92.98 & 10.25 & 98.06 & 63.80 & 87.10 & 63.76 & 87.21 & \underline{44.10} & \underline{90.71} & 78.48 & 78.00 & \underline{50.61} & \underline{89.01} \\
& \textbf{Ours} & \textbf{28.21} & \textbf{95.32} & 14.41 & 97.29 & 64.16 & 85.74 & 63.54 & 85.89 & \textbf{31.91} & \textbf{93.27} & \underline{77.81} & 77.59 & \textbf{46.67} & \textbf{89.18} \\

\midrule
\multirow{10}{*}{\begin{tabular}[c]{@{}c@{}}CIFAR-100 \\DenseNet\end{tabular}} 
& MSP~\cite{MSP} & 82.03 & 75.19 & 60.54 & 85.60 & 85.30 & 69.17 & 86.05 & 70.17 & 85.14 & 71.21 & 82.15 & 74.78 & 80.20 & 74.35 \\
& ODIN~\cite{ODIN} & 45.14 & 91.75 & 9.54 & 98.20 & 59.14 & 86.34 & 61.46 & 85.87 & 81.72 & 72.21 & 80.38 & \underline{77.90} & 56.23 & 85.38 \\
& Mahalanobis~\cite{mahala} & 52.30 & 89.16 & 86.21 & 68.78 & 30.47 & 94.19 & 30.22 & 93.81 & \underline{34.10} & \underline{89.52} & 94.79 & 57.42 & 54.68 & 82.15 \\
& Energy~\cite{energy} & 88.04 & 81.30 & 14.78 & 97.43 & 70.72 & 80.14 & 74.60 & 78.95 & 85.16 & 70.98 & \textbf{77.93} & \textbf{78.25} & 68.54 & 81.18 \\
& BATS~\cite{bats} & 91.80 & 76.53 & 21.19 & 95.55 & 41.56 & 92.87 & 44.10 & 92.27 & 63.17 & 82.02 & \underline{79.29} & 75.65 & 56.85 & 85.82 \\
& DICE~\cite{dice} & 60.07 & 88.20 & \textbf{0.93} & \textbf{99.74} & 51.76 & 89.32 & 49.59 & 89.51 & 61.24 & 77.22 & 80.20 & 77.65 & 50.63 & 86.94 \\
& ReAct~\cite{react} & 82.70 & 82.22 & 19.49 & 96.33 & 63.85 & 85.98 & 68.11 & 84.96 & 78.33 & 79.08 & 79.76 & 76.21 & 65.37 & 84.13 \\
& DICE+ReAct~\cite{dice} & 55.54 & 88.03 & 7.57 & 98.61 & 54.44 & 89.84 & 44.32 & 91.45 & 41.29 & 86.11 & 93.75 & 56.48 & 49.49 & 85.09 \\
& LINe~\cite{LINe} & \underline{31.13} & \underline{91.89} & 5.76 & 98.85 & \textbf{25.36} & \underline{94.54} & \textbf{24.15} & \underline{94.76} & 39.18 & 87.85 & 88.44 & 64.17 & \underline{35.67} & \underline{88.68} \\
& \textbf{Ours} & \textbf{14.04} & \textbf{96.90} & \underline{3.27} & \underline{99.28} & \underline{29.97} & \textbf{94.89} & \underline{25.25} & \textbf{95.51} & \textbf{25.05} & \textbf{93.90} & 82.78 & 72.87 & \textbf{30.06} & \textbf{92.23} \\

\bottomrule

\end{tabular}%
}
\end{center}

\end{table*}

\subsection{Experimental Setup}
\subsubsection{Datasets}
Our method is extensively evaluated on the widely used CIFAR benchmarks~\cite{cifar} and the large-scale OOD detection benchmark based on ImageNet~\cite{mos}. 
For CIFAR benchmarks, CIFAR-10 and CIFAR-100~\cite{cifar} are respectively used as in-distribution datasets, with 50,000 training images and 10,000 test images per dataset. Six OOD datasets are used during testing, including SVHN~\cite{SVHN}, LSUN-Crop~\cite{yu2016lsun}, LSUN-Resize~\cite{yu2016lsun}, iSUN~\cite{isun}, Textures~\cite{textures}, and Places365~\cite{zhou2017places}. For the ImageNet benchmark,  ImageNet-1k~\cite{deng2009imagenet} is used as the in-distribution dataset and four OOD datasets are used during testing, including Places365~\cite{zhou2017places}, Textures~\cite{textures}, iNaturalist~\cite{van2018inaturalist}, and SUN~\cite{sun}.

\subsubsection{Implementation Details}
Following the common experimental setting~\cite{dice, react, mos}, ResNet-18~\cite{resnet} and DenseNet~\cite{densenet} are used as the backbones on CIFAR benchmarks. During model training, each CIFAR image is randomly cropped to 32 $\times$ 32 pixels and then randomly flipped horizontally. The models are trained with batch size 128 for 100 epochs, weight decay 0.0001, and momentum 0.9. The initial learning rate is 0.1 and decays by a factor of 10 at epochs 50, 75, and 90. On the ImageNet benchmark, the pre-trained ResNet-50~\cite{resnet} and MobileNetV2~\cite{sandler2018mobilenetv2} provided by Pytorch are adopted. During testing, all images are resized to 32 $\times$ 32 pixels on CIFAR benchmarks, and to 256 $\times$ 256 and center crop to size of 224 $\times$ 224 pixels on the ImageNet benchmark. All training images are used to compute class prototypes for each model on both CIFAR and ImageNet benchmarks. 

{\subsubsection{Evaluation Metrics}
We measure the performance of OOD detection using the two most widely evaluation metrics:~(1) FPR95: the false positive rate when the true positive rate is 95\%. Lower FPR95 means better OOD detection performance and vice versa.~(2) AUROC: the area under the receiver operating characteristic curve. Larger AUROC means better performance.}
\subsubsection{Competitive methods for comparison}
Our method is compared with multiple competitive post-hoc OOD detection methods, including MSP~\cite{MSP}, ODIN~\cite{ODIN}, Mahalanobis~\cite{mahala}, Energy~\cite{energy}, BATS~\cite{bats}, DICE~\cite{dice}, ReAct~\cite{react}, DICE+ReAct~\cite{dice}, and LINe~\cite{LINe}.

\subsection{Evaluation on CIFAR Benchmarks}
Table~\ref{tab:cifar benchmark} summarizes the comparisons between our method and competitive post-hoc OOD detection methods on CIFAR-10 and CIFAR-100 benchmarks. Our method achieves state-of-the-art performance with both ResNet-18 and DenseNet backbones. On the CIFAR-10 benchmark, our method with DenseNet outperforms the strongest baseline by 2.33\% in FPR95.
On the CIFAR-100 benchmark, our method with DenseNet outperforms the competitive method ReAct~\cite{react} by 35.31\% in FPR95 and 8.1\% in AUROC.
Compared to the state-of-the-art method LINe~\cite{LINe}, our method reduces FPR95 by 5.61\% and improves AUROC by 3.55\% on DenseNet while reducing FPR95 by 3.94\% on ResNet-18. These results consistently support the effectiveness of our method on different model backbones for OOD detection.

\subsection{Evaluation on ImageNet Benchmark}
Table~\ref{tab:imagenet1k benchmark} shows the OOD detection performance of our method and competitive baselines with ResNet-50 and MobileNetV2 backbones. The detailed performance on four datasets and the average over the four datasets are reported. It shows that our method achieves state-of-the-art performance on average with both backbones. For example, with ResNet-50, our method outperforms Energy~\cite{energy}  by 38.79\% in FPR95 and 9.36\% in AUROC. Our method reduces FPR95 by 11.81\% compared to ReAct~\cite{react}, which confirms the importance of feature masking and logit smoothing for OOD detection. Also, our method outperforms DICE+ReAct~\cite{dice} and LINe~\cite{LINe} on ResNet-50 by 7.63\% and 1.08\% in FPR95, respectively. This further confirms the superiority of our method particularly considering that the two methods and ours all use ReAct to improve performance. 
Note that our method does not achieve the best performance on some individual OOD datasets like SUN and Places with the ResNet-50 backbone, probably because (OOD) images in SUN and Places are similar to those (ID) images in ImageNet in the feature space {(see in the Figure~\ref{fig:cosine})} such that logit smoothing is hard to further improves OOD detection performance.

\begin{table*}[!tbh]
\centering
\caption{Performance comparison on the ImageNet benchmark with ResNet-50 and MobileNet backbones. 
 }
\label{tab:imagenet1k benchmark}
\resizebox{\textwidth}{!}{%
\begin{tabular}{cccccccccccc}
\toprule
\multicolumn{1}{c}{\multirow{3}{*}{\begin{tabular}[c]{@{}c@{}} \textbf{Model}\end{tabular}}} & \multicolumn{1}{c}{\multirow{3}{*}{\textbf{Method}}} & \multicolumn{8}{c}{\textbf{OOD Datasets}} & \multicolumn{2}{c}{\multirow{2}{*}{\textbf{Average}}}\\ \cline{3-10} 
\multicolumn{1}{c}{} & \multicolumn{1}{c}{} & \multicolumn{2}{c}{\textbf{iNaturalist}} & \multicolumn{2}{c}{\textbf{SUN}} & \multicolumn{2}{c}{\textbf{Places}} & \multicolumn{2}{c}{\textbf{Textures}}  \\
\multicolumn{1}{c}{} & \multicolumn{1}{c}{} & \multicolumn{1}{c}{FPR95$\downarrow$} & \multicolumn{1}{c}{AUROC$\uparrow$} & \multicolumn{1}{c}{FPR95$\downarrow$} & \multicolumn{1}{c}{AUROC$\uparrow$} & \multicolumn{1}{c}{FPR95$\downarrow$} & \multicolumn{1}{c}{AUROC$\uparrow$} & \multicolumn{1}{c}{FPR95$\downarrow$} & \multicolumn{1}{c}{AUROC$\uparrow$} & \multicolumn{1}{c}{FPR95$\downarrow$} & \multicolumn{1}{c}{AUROC$\uparrow$} \\ \midrule

\multirow{10}{*}{\begin{tabular}[c]{@{}c@{}} ResNet-50\end{tabular}} 
& MSP~\cite{MSP} & 54.99 & 87.74 & 70.83 & 80.86 & 73.99 & 79.76 & 68.00 & 79.61 & 66.95 & 81.99 \\
& ODIN~\cite{ODIN} & 47.66 & 89.66 & 60.15 & 84.59 & 67.89 & 81.78 & 50.23 & 85.62 & 56.48 & 85.41 \\
& Mahalanobis~\cite{mahala} & 97.00 & 52.65 & 98.50 & 42.41 & 98.40 & 41.79 & 55.80 & 85.01 & 87.43 & 55.47 \\
& Energy~\cite{energy} & 55.72 & 89.95 & 59.26 & 85.89 & 64.92 & 82.86 & 53.72 & 85.99 & 58.41 & 86.17 \\
& BATS~\cite{bats} & 12.57 & \underline{97.67} & \underline{22.62} & \textbf{95.33} & 34.34 & 91.83 & 38.90 & 92.27 & 27.11 & 94.28 \\
& DICE~\cite{dice} & 25.63 & 94.49 & 35.15 & 90.83 & 46.49 & 87.48 & 31.72 & 90.30 & 34.75 & 90.77 \\
& ReAct~\cite{react} & 20.38 & 96.22 & 24.20 & 94.20 & 33.85 & 91.58 & 47.30 & 89.80 & 31.43 & 92.95 \\
& DICE + ReAct~\cite{dice} & 18.64 & 96.24 & 25.45 & 93.94 & 36.86 & 90.67 & 28.07 & 92.74 & 27.25 & 93.40 \\
& LINe~\cite{LINe} & \underline{12.26} & 97.56 & \textbf{19.48} & \underline{95.26} & \textbf{28.52} & \textbf{92.85} & \underline{22.54} & \underline{94.44} & \underline{20.70} & \underline{95.03} \\
& Ours & \textbf{8.91} & \textbf{98.18} & 23.08 & 94.78 & \underline{32.50} & \underline{92.17} & \textbf{13.99} & \textbf{96.97} & \textbf{19.62} & \textbf{95.53} \\

\midrule

\multirow{10}{*}{\begin{tabular}[c]{@{}c@{}}MobileNet\end{tabular}} 
& MSP~\cite{MSP} & 64.29 & 85.32 & 77.02 & 77.10 & 79.23 & 76.27 & 73.51 & 77.30 & 73.51 & 79.00 \\
& ODIN~\cite{ODIN} & 55.39 & 87.62 & 54.07 & 85.88 & 57.36 & 84.71 & 49.96 & 85.03 & 54.20 & 85.81 \\
& Mahalanobis~\cite{mahala} & 62.11 & 81.00 & 47.82 & 86.33 & 52.09 & 83.63 & 92.38 & 33.06 & 63.60 & 71.01 \\
& Energy~\cite{energy} & 59.50 & 88.91 & 62.65 & 84.50 & 69.37 & 81.19 & 58.05 & 85.03 & 62.39 & 84.91 \\
& BATS~\cite{bats} & 31.56 & 94.33 & 41.68 & 90.21 & 52.43 & 86.26 & 38.69 & 90.76 & 41.09 & 90.39 \\
& DICE~\cite{dice} & 43.09 & 90.83 & 38.69 & 90.46 & 53.11 & 85.81 & 32.80 & 91.30 & 41.92 & 89.60 \\
& ReAct~\cite{react} & 42.40 & 91.53 & 47.69 & 88.16 & 51.56 & 86.64 & 38.42 & 91.53 & 45.02 & 89.47 \\
& DICE + ReAct~\cite{dice} & 32.30 & 93.57 & \textbf{31.22} & 92.86 & \underline{46.78} & 88.02 & 16.28 & 96.25 & 31.64 & 92.68 \\
& LINe~\cite{LINe} & \underline{24.95} & \underline{95.53} & \underline{33.19} & \underline{92.94} & 47.95 & \underline{88.98} & \textbf{12.30} & \textbf{97.05} & \underline{29.60} & \underline{93.62} \\
& Ours & \textbf{20.43} & \textbf{96.41} & 33.51 & \textbf{93.32} & \textbf{45.64} & \textbf{89.97} & \underline{15.28} & \underline{96.94} & \textbf{28.72} & \textbf{94.16} \\
 \bottomrule

\end{tabular}%
}

\end{table*}

\begin{table}[!tbh]
\centering
\caption{Ablation study of different components in our method. DenseNet is used on the CIFAR-100 benchmark, and ResNet-50 on the ImageNet benchmark. 'FM' and 'LS' denote feature masking and logit smoothing respectively. All percentage values are averaged over multiple OOD datasets. }
\label{Ablation FP LS}
\resizebox{1.0\linewidth}{!}{
\begin{tabular}{ccc|cc}
\toprule

\multicolumn{1}{c}{\multirow{2}{*}{\begin{tabular}[c]{@{}c@{}} \textbf{ReAct}\end{tabular}}} & 
\multicolumn{1}{c}{\multirow{2}{*}{\textbf{FM}}} & 
\multicolumn{1}{c|}{\multirow{2}{*}{\textbf{LS}}} & 
\textbf{CIFAR-100} & \textbf{ImageNet} \\ 
\multicolumn{1}{c}{} & \multicolumn{1}{c}{} & \multicolumn{1}{c|}{} & \textbf{FPR95$\downarrow$ / AUROC$\uparrow$} & \textbf{FPR95$\downarrow$ / AUROC$\uparrow$}  \\
\midrule
 &  &  & 68.54 / 81.18 & 58.41 / 86.17 \\
\checkmark &  &  & 65.37 / 84.13 & 31.43 / 92.95 \\
\checkmark & \checkmark &  & 36.74 / 90.01 & 26.33 / 93.38 \\
\checkmark &  & \checkmark & 48.42 / 88.77 & 23.10 / 95.09 \\
 & \checkmark & \checkmark & 40.34 / 89.89 & 32.51 / 92.36 \\
\checkmark & \checkmark & \checkmark & \textbf{30.06} / \textbf{92.23} & \textbf{19.62} / \textbf{95.53} \\
\bottomrule
\end{tabular}
}

\end{table}

\subsection{Ablation and Sensitivity Studies}

\subsubsection{Ablation of different components in our method}
Our method includes feature masking ('FM') and logit smoothing ('LS') as well as the ReAct clipping~\cite{react} to improve OOD detection performance. Table~\ref{Ablation FP LS} shows an ablation study of each component in our method. 
Compared to ReAct~\cite{react} only (second row), ReAct+FM (third row) reduces FPR95 by 28.63\% on the CIFAR-100 benchmark and reduces FPR95 by 5.1\% on the ImageNet benchmark, which supports the effect of feature masking on performance improvement. ReAct+LS (fourth row) outperforms ReAct by 16.95\% and 8.33\% in FPR95 on the CIFAR-100 benchmark and the ImageNet benchmark respectively, supporting the effectiveness of logit smoothing for OOD detection. Besides, when we remove ReAct in our method (fifth row), the FPR95 increases by 10.28\% and 12.89\% compared to our method (last row) on the CIFAR-100 benchmark and the ImageNet benchmark respectively, suggesting that ReAct is also important for performance improvement. The inclusion of all these three components (last row) achieves the best OOD detection performance on both benchmarks, supporting that the three components are complementary to each other and all play important roles in our method for performance improvement.



\begin{figure}[!tbh]
    \centering
    \includegraphics[width=0.48\textwidth]{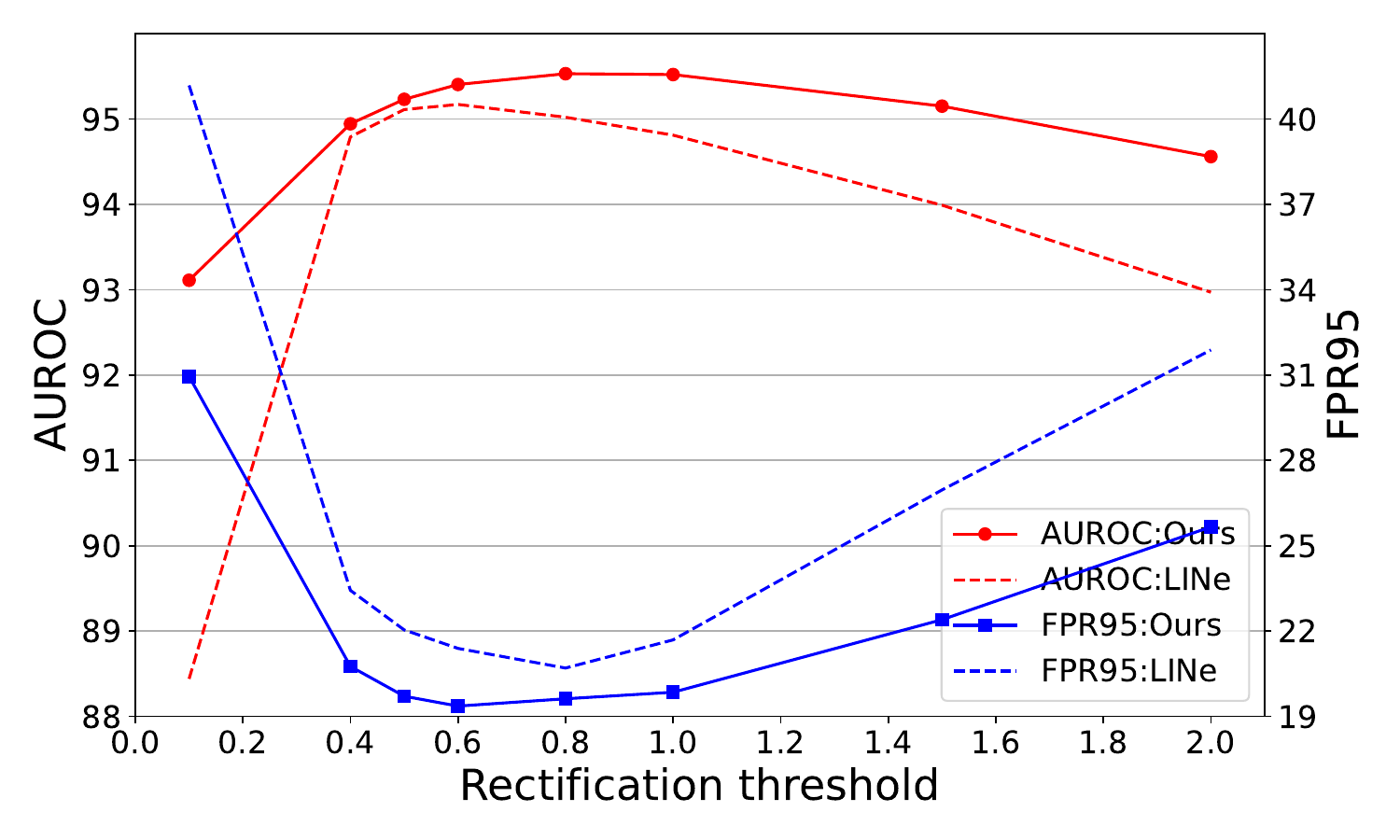}
    \caption{Sensitivity study of rectification thresholds $\lambda$ on the ImageNet benchmark with ResNet-50. 
    All values are percentages and averaged over multiple OOD datasets.  }
    \label{fig:Threshold Ablation}
\end{figure}
\subsubsection{Sensitivity of rectification threshold}
As ReAct is one of the important components in our method, we perform a sensitivity study to show the effect of rectification threshold $\lambda$ and compare the performance between our method and LINe~\cite{LINe} which also uses ReAct. As shown in Figure~\ref{fig:Threshold Ablation}, 
when the threshold $\lambda$ is too large (e.g., 2.0), both methods perform relatively worse because ReAct plays little role in clipping large feature activation. As the threshold $\lambda$ decreases, the performance improves and our method always outperforms the state-of-the-art method LINe for the same rectification threshold $\lambda$. Our method performs stably well in the range $[0.5, 1.5]$, suggesting that our method is robust to the choice of the hyperparameter $\lambda$.
The performance of both methods drops when $\lambda$ approaches 0, because most of the feature activation values are rectified to a small threshold which in turn leads to poorer logit separability between test ID and OOD images.

\subsubsection{Sensitivity of masking percentile}

\begin{figure*}[!tbh]
    \centering
    \subfloat[CIFAR-10]{\includegraphics[width=0.33\textwidth]{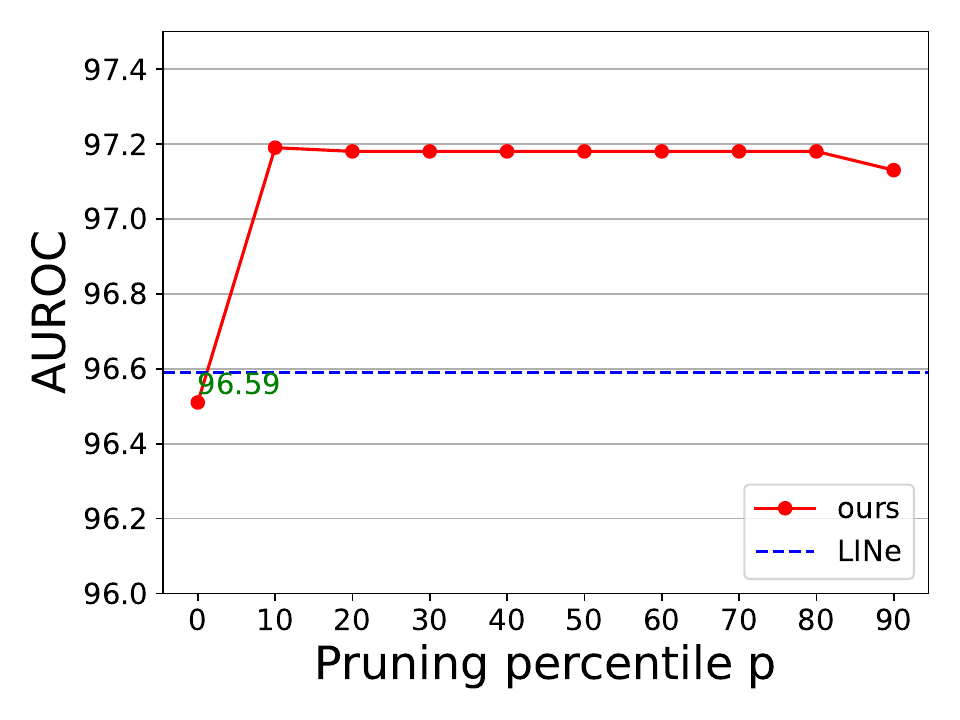}%
    \label{fig:/CIFAR-10_ours}}
    \hfil
    \subfloat[CIFAR-100]{\includegraphics[width=0.33\textwidth]{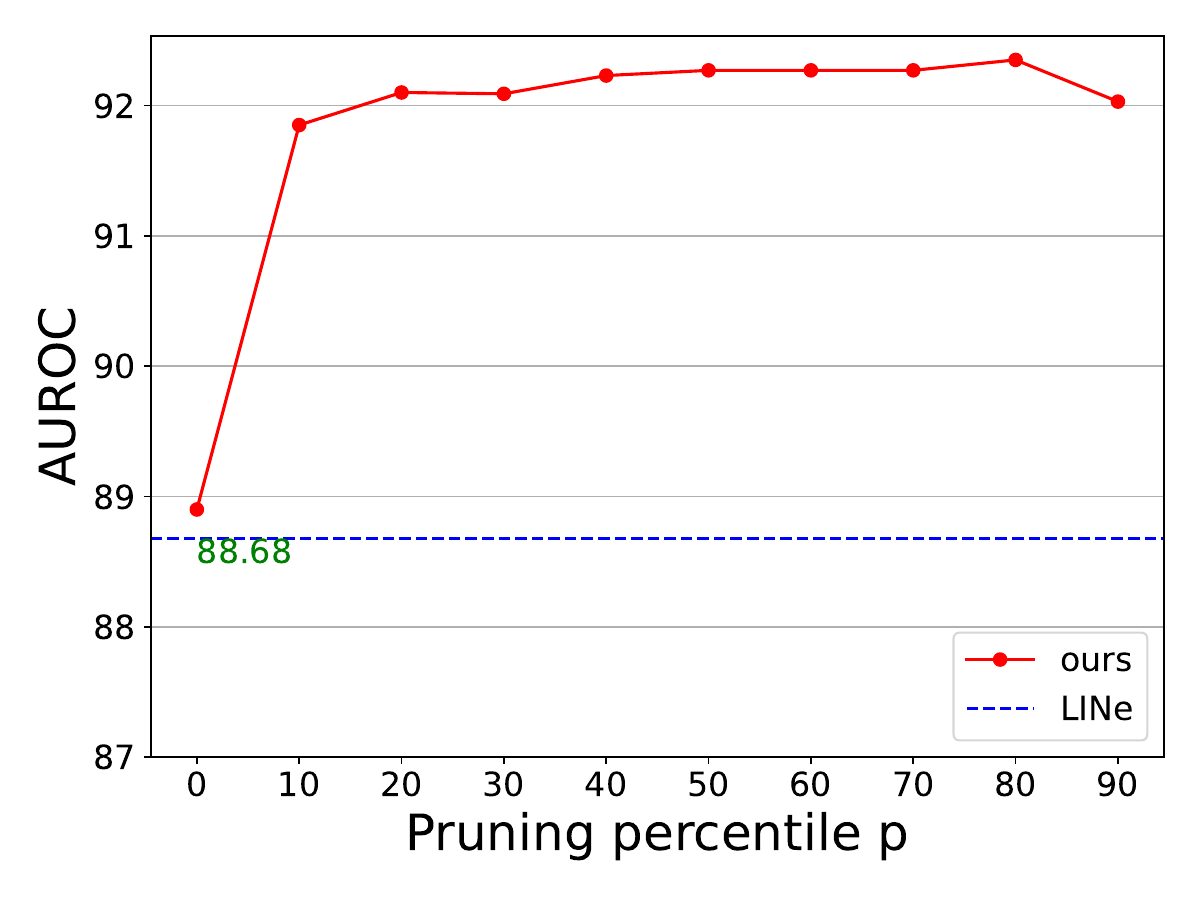}%
    \label{fig:/CIFAR-100_ours}}
    \hfil
    \subfloat[ImageNet]{\includegraphics[width=0.33\textwidth]{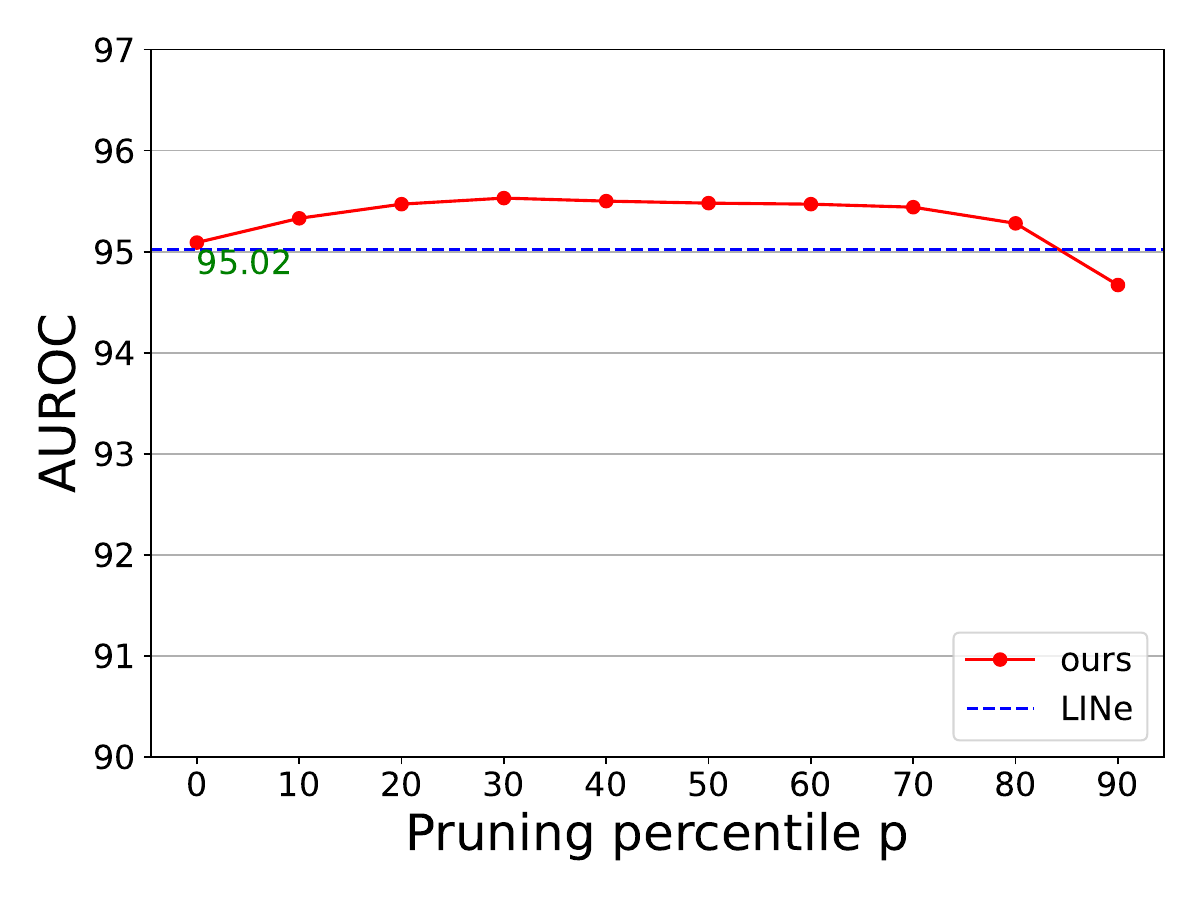}%
    \label{fig:/ImageNet_ours}}
    \hfil
    \caption{{Sensitivity study of masking percentile $p$ on the CIFAR-10, CIFAR-100, and ImageNet benchmarks.  
    DenseNet is used on CIFAR benchmarks, and ResNet-50 on the ImageNet benchmark.}
    All AUROC values are averaged over multiple OOD datasets.}
    \label{fig:ablation percentile}
\end{figure*}

Since feature masking is an important part of our method, we also perform a sensitivity study of masking percentile $p = \frac{L-k}{L} \cdot 100\%$, where $L$ is the total number of elements in the feature vector at the penultimate layer and $k$ is the number of feature elements selected for un-masking. 
Figure~\ref{fig:ablation percentile} demonstrates the performance of our method (red curves) on the CIFAR-10 and CIFAR-100 benchmarks with DenseNet and on the ImageNet benchmark with ResNet-50 when varying the masking percentile $p$. The performance from the state-of-the-art method LINe~\cite{LINe} is also included for comparison. 
Note that $p=0$ corresponds to the case of using the original feature vector (i..e, no feature masking). 
Significant performance improvement is observed when varying $p$ from $0\%$ to $10\%$, clearly supporting the importance of feature masking for OOD detection. The performance of our method remains stably well between the large range $[10\%, 80\%]$  and is consistently better than that of the strong baseline LINe on both benchmarks, confirming that our method is insensitive to the choice of hyper-parameter in feature masking. When $p$ gets extremely large (e.g., $90\%$), the performance drops rapidly on the ImageNet benchmark as expected, because multiple feature elements which are crucial to ID classes have been masked at such high masking percentile.  




\subsection{Compatibility with other methods}

We have shown the effectiveness of applying our method to Energy score~\cite{energy}. Actually, our method is also compatible with other existing methods. In Table~\ref{different method}, we compare the results of applying our method and ReAct~\cite{react} to different OOD detection methods, including MSP~\cite{MSP}, ODIN~\cite{ODIN}, LINe~\cite{LINe}, and Energy~\cite{energy}. Note that MSP, ODIN, and Energy are associated with different scoring functions, therefore our method and ReAct can simply be applied to these methods. As for LINe, we only apply logit smoothing to it, since it already masks features and applies ReAct.  
As we can see, our method can improve the performance of all these methods and outperform all the methods with ReAct, which confirms the flexibility and effectiveness of our method. 
Notably, when applying logit smoothing to LINe, the best performance is achieved on the ImageNet benchmark with ResNet-50, further confirming the importance of logit smoothing in OOD detection. 

\begin{table}[!tbh]
\centering
\caption{Results of applying our method and ReAct to different OOD detection methods. DenseNet is used on CIFAR benchmarks, and ResNet-50 on the ImageNet benchmark. 
All values are averaged over multiple OOD datasets. }
\label{different method}
\resizebox{1.0\linewidth}{!}{

\begin{tabular}{ccccccc}
\toprule
\multirow{2}{*}{\textbf{Method}} & \multicolumn{2}{c}{\textbf{CIFAR-10}} & \multicolumn{2}{c}{\textbf{CIFAR-100}} & \multicolumn{2}{c}{\textbf{ImageNet}} \\
&\textbf{FPR95}$\downarrow$ &\textbf{AUROC}$\uparrow$ &\textbf{FPR95}$\downarrow$ &\textbf{AUROC}$\uparrow$ &\textbf{FPR95}$\downarrow$ &\textbf{AUROC}$\uparrow$ \\

\midrule
MSP & 50.04 & 92.05 & 80.20 & 74.35 & 66.95 & 81.99\\
MSP+ReAct & 50.07 & 92.32 & 80.19 & 74.71 & 58.28 & 87.06\\
MSP+Ours & \textbf{18.99} & \textbf{96.62} & \textbf{43.49} & \textbf{89.25} & \textbf{22.49} & \textbf{94.66} \\
\midrule
Energy & 28.28 & 94.31 & 68.54 & 81.18 & 58.41 & 86.17\\
Energy+ReAct & 23.47 & 95.95 & 65.37 & 84.13 & 31.43 & 92.95\\
Energy+Ours & \textbf{14.62} & \textbf{97.18} & \textbf{30.06} & \textbf{92.23} & \textbf{19.62 }& \textbf{95.53} \\
\midrule
ODIN & 22.14 & 94.17 & 56.23 & 85.38 & 56.48 & 85.41  \\
ODIN+ReAct & 22.08 & 95.63 & 49.18 & 88.85 & 44.10 & 90.70  \\
ODIN+Ours & \textbf{14.60} & \textbf{97.10} & \textbf{28.42} & \textbf{92.15} & \textbf{21.16} & \textbf{95.05} \\
\midrule
LINe & 16.95 & 96.59 & 35.67 & 88.68 & 20.70 & 95.03  \\
LINe+Ours & \textbf{16.35} & \textbf{96.71} & \textbf{33.37} & \textbf{89.43} & \textbf{17.84} & \textbf{95.97} \\
\bottomrule
\end{tabular}
}

\end{table}

{\subsection{Validation Strategy}
We use a validation set of Gaussian noise images generated from $\mathcal{N}$(0,1) for each pixel.
We select the optimal masking percentile $p$ from $\{$10, 20, 30, 40, 50, 60, 70, 80, 90$\}$. 
We choose hyperparameters $p=60\%$ for CIFAR-10 and CIFAR-100 datasets, and choose $p=30\%$ for the ImageNet dataset.
As for rectification threshold $\lambda$, we choose $\lambda=1.6$ for CIFAR-10 and CIFAR-100 datasets with DenseNet, and $\lambda=1.0$ with ResNet-18. 
For the ImageNet dataset, we choose $\lambda=0.8$ with ResNet-50, and $\lambda=0.2$ with MobileNet. Note that the performance of our method is largely insensitive to the choice of these hyper-parameters, as shown in Fig\ref{fig:ablation percentile}.

\subsection{Evaluation on Transformer-based ViT Model}
Following the general experimental setting, we have shown the effectiveness of our method on various CNN-based models. 
To further explore the generalizability of our method to more model architectures, we provide a comprehensive evaluation on the ImageNet benchmark with transformer-based ViT Model~\cite{vit}. ViT~\cite{vit} is a transformer-based image classification model which treats images as sequences of patches. We use the ViT-B/16 model which is pre-trained on ImageNet-21K and fine-tuned on ImageNet-1K.

In Table~\ref{tab:imagenet1k vit}, we report the performance of OOD detection between various post-hoc detection methods.
It indicates that our method achieves the best performance on average with ViT backbone.
Note that transformer-based models are quite different from CNN-based models, the observations on CNN-based models do not necessarily apply to ViT models. For example, the performance of DICE~\cite{dice} and  ReAct~\cite{react} drop compared to Energy~\cite{energy}.
However, our method outperforms Energy by 1.2\% in FPR95 on average which demonstrates the effectiveness and generalizability of our method with ViT backbone.}

\begin{table*}[!tbh]
\centering
\caption{Performance comparison on the ImageNet benchmark with ViT-B/16 model. 
 }
\label{tab:imagenet1k vit}
\resizebox{\textwidth}{!}{%
\begin{tabular}{ccccccccccc}
\toprule
\multicolumn{1}{c}{\multirow{3}{*}{\textbf{Method}}} & \multicolumn{8}{c}{\textbf{OOD Datasets}} & \multicolumn{2}{c}{\multirow{2}{*}{\textbf{Average}}}\\ \cline{2-9} 
\multicolumn{1}{c}{} & \multicolumn{2}{c}{\textbf{iNaturalist}} & \multicolumn{2}{c}{\textbf{SUN}} & \multicolumn{2}{c}{\textbf{Places}} & \multicolumn{2}{c}{\textbf{Textures}}  \\
\multicolumn{1}{c}{} & \multicolumn{1}{c}{FPR95$\downarrow$} & \multicolumn{1}{c}{AUROC$\uparrow$} & \multicolumn{1}{c}{FPR95$\downarrow$} & \multicolumn{1}{c}{AUROC$\uparrow$} & \multicolumn{1}{c}{FPR95$\downarrow$} & \multicolumn{1}{c}{AUROC$\uparrow$} & \multicolumn{1}{c}{FPR95$\downarrow$} & \multicolumn{1}{c}{AUROC$\uparrow$} & \multicolumn{1}{c}{FPR95$\downarrow$} & \multicolumn{1}{c}{AUROC$\uparrow$} \\ \midrule

MSP~\cite{MSP} & 19.15 & 96.13 & 57.00 & 86.13 & 59.97 & 85.09 & 51.49 & 85.10 & 46.90 & 88.11 \\
ODIN~\cite{ODIN} & 6.53 & 98.57 & 38.99 & 91.50 & 46.66 & 89.24 & 33.99 & 91.29 & 31.54 & 92.65 \\
Mahalanobis~\cite{mahala} & \textbf{2.13} & \textbf{99.54} & 51.26 & 89.14 & 59.87 & 86.24 & \underline{28.32} & \underline{92.60} & 35.39 & 91.88 \\
Energy~\cite{energy} & 6.04 & 98.66 & \underline{37.11} & 91.75 & \underline{45.30} & 89.30 & 31.90 & 91.70 & \underline{30.09} & 92.85 \\
BATS~\cite{bats} & 6.23 & 98.59 & 41.51 & 90.41 & 49.62 & 87.85 & 33.16 & 91.23 & 32.63 & 92.02 \\
DICE~\cite{dice} & 91.86 & 65.22 & 88.70 & 64.47 & 93.85 & 61.53 & 74.27 & 74.95 & 87.17 & 66.54 \\
ReAct~\cite{react} & 4.19 & 99.01 & 39.06 & 91.54 & 47.38 & 89.11 & 32.27 & 91.54 & 30.72 & 92.80 \\
DICE + ReAct~\cite{dice} & 96.13 & 55.25 & 89.42 & 63.32 & 95.03 & 59.73 & 78.21 & 71.03 & 89.70 & 62.33 \\

LINe~\cite{LINe} & 4.36 & 98.93 & \textbf{33.79} & \textbf{93.28} & \textbf{45.09} & \textbf{90.62} & 47.46 & 88.63 & 32.68 & \underline{92.87} \\
Ours & \underline{3.00} & \underline{99.12} & 37.56 & \underline{92.25} & 46.70 & \underline{89.63} & \textbf{28.30} & \textbf{93.15} & \textbf{28.89} & \textbf{93.54} \\

\bottomrule

\end{tabular}%
}

\end{table*}

\section{Conclusion}
In this study, a new post-hoc OOD detection method is proposed based on feature masking and logit smoothing. 
Feature masking is expected to remove those high activation features caused by OOD samples, while preserving most of the high activation features caused by ID samples. Logit smoothing can further enlarge the difference in OOD score between ID and OOD samples, therefore helping improve OOD detection performance. 
Extensive experiments confirm that our method establishes new state-of-the-art performance on multiple benchmarks with different model and is robust to the choice of hyperparameters. 
Moreover, the flexibility combination of our method with existing OOD detection methods suggests the high extensibility of our method. 
{We provide a novel insight for OOD detection by combining the information in the feature space with logit and hope that our work can help to motivate new research on solving the overconfidence problem of neural networks by using different information in the feature space.}

\bibliography{IEEE}
\bibliographystyle{IEEEtran}



\vfill

\end{document}